\documentclass[journal]{IEEEtran}

\usepackage{cite}
\usepackage{amsmath, amssymb, amsfonts}
\usepackage{algorithm, algpseudocode}
\usepackage{graphicx, epsfig}
\usepackage{textcomp}
\usepackage{url}
\usepackage{booktabs}
\usepackage{subfigure}
\usepackage{amsthm}
\usepackage{times}
\usepackage{xcolor}
\usepackage{soul}
\usepackage[colorlinks, linkcolor=blue]{hyperref}
\usepackage{lipsum}   
\usepackage{tikz}  

\usepackage{graphicx}
\usepackage{booktabs}

\usepackage{pifont}

\usepackage{amssymb} 
\usepackage{pifont}  
\usepackage{colortbl} 

\def\BibTeX{{\rm B\kern-.05em{\sc i\kern-.025em b}\kern-.08em
    T\kern-.1667em\lower.7ex\hbox{E}\kern-.125emX}}

\begin{document}

\title{Seeing Isn’t Always Believing: Analysis of Grad-CAM Faithfulness and Localization Reliability in Lung Cancer CT Classification}

\author{Teerapong~Panboonyuen%
\thanks{T. Panboonyuen (also known as Kao) is with Chulalongkorn University, Bangkok 10330, Thailand (e-mail: teerapong.panboonyuen@gmail.com). \\ The project is at: \url{https://kaopanboonyuen.github.io/GradFaith-CAM/}. \\ ORCID: \href{https://orcid.org/0000-0001-8464-4476}{0000-0001-8464-4476}}%
\thanks{This work is motivated by a deep commitment to making artificial intelligence
not only accurate, but genuinely trustworthy and interpretable. In high-stakes
applications such as lung cancer CT analysis, where decisions directly affect
human lives, it is essential that AI systems truly understand relevant medical
evidence rather than merely achieving high predictive performance. This study
aims to promote the responsible and correct use of AI by rigorously examining
the faithfulness and localization reliability of explainable methods, with a
particular focus on Grad-CAM.}

\vspace{2mm}
\small \url{https://kaopanboonyuen.github.io/GradFaith-CAM/} \\


}

\markboth{Accepted for publication in KST2026: 18th International Conference on Knowledge and Smart Technology}%
{Seeing Isn’t Always Believing: Grad-CAM Faithfulness and Localization Reliability}

\maketitle

\begin{abstract}
Explainable Artificial Intelligence (XAI) techniques, such as Gradient-weighted Class Activation Mapping (Grad-CAM), have become indispensable for visualizing the reasoning process of deep neural networks in medical image analysis. Despite their popularity, the faithfulness and reliability of these heatmap-based explanations remain under scrutiny. This study critically investigates whether Grad-CAM truly represents the internal decision-making of deep models trained for lung cancer image classification. Using the publicly available \textit{IQ-OTH/NCCD} dataset, we evaluate five representative architectures — \textbf{ResNet-50}, \textbf{ResNet-101}, \textbf{DenseNet-161}, \textbf{EfficientNet-B0}, and \textbf{ViT-Base-Patch16-224} — to explore model-dependent variations in Grad-CAM interpretability. We introduce a quantitative evaluation framework that combines localization accuracy, perturbation-based faithfulness, and explanation consistency to assess Grad-CAM reliability across architectures. Experimental findings reveal that while Grad-CAM effectively highlights salient tumor regions in most convolutional networks, its interpretive fidelity significantly degrades for Vision Transformer models due to non-local attention behavior. Furthermore, cross-model comparison indicates substantial variability in saliency localization, implying that Grad-CAM explanations may not always correspond to the true diagnostic evidence used by the networks. This work exposes critical limitations of current saliency-based XAI approaches in medical imaging and emphasizes the need for model-aware interpretability methods that are both computationally sound and clinically meaningful. Our findings aim to inspire a more cautious and rigorous adoption of visual explanation tools in medical AI, urging the community to rethink what it truly means to “trust” a model’s explanation.
\end{abstract}

\begin{tikzpicture}[remember picture,overlay]
\node[anchor=south,yshift=5pt] at (current page.south) {\footnotesize
\begin{minipage}{0.95\textwidth}
\centering
``This is the author’s version of the paper accepted for publication in the
\emph{KST2026}.\\
This manuscript is shared for academic use and is also available as a preprint on arXiv.
The final published version may differ from this version."
\end{minipage}
};
\end{tikzpicture}


\section{Introduction}
\IEEEPARstart{A}{ccurate} classification of lung cancer from CT images is critical for early diagnosis and effective treatment planning. While deep learning approaches, particularly convolutional neural networks (CNNs) and vision transformers, have shown remarkable predictive capabilities, there remains a fundamental challenge: understanding \emph{why} a model makes a particular decision. In high-stakes applications such as oncology, interpretability and faithfulness of model explanations are as important as raw accuracy. This work is motivated by the premise expressed in our title, \textit{``Seeing Isn’t Always Believing: Grad-CAM Faithfulness Analysis on Lung Cancer Image Classification''}, emphasizing that high confidence predictions do not necessarily guarantee clinically reliable localization or understanding of malignant regions.

We utilize the Iraq-Oncology Teaching Hospital/National Center for Cancer Diseases (IQ-OTH/NCCD) lung cancer dataset, which consists of 1190 CT slices from 110 patients collected over three months in fall 2019. The dataset covers three clinically relevant classes: \textbf{Normal}, \textbf{Benign}, and \textbf{Malignant}, each carefully annotated by expert oncologists and radiologists. CT acquisition protocols included 120 kV scans, 1 mm slice thickness, window widths from 350 to 1200 HU, and window centers from 50 to 600 HU, all captured at full inspiration. Every image was de-identified in compliance with ethical standards, and written consent was waived by the institutional review boards of the participating hospitals. These scans represent a diverse cohort, encompassing variations in age, gender, and other demographics, thereby providing a realistic testbed for evaluating model generalization.

In this study, we evaluate and compare five state-of-the-art deep learning models: \textbf{ResNet50}, \textbf{ResNet101}, \textbf{DenseNet161}, \textbf{EfficientNetB0}, and \textbf{ViT-Base-Patch16-224}. Our analysis goes beyond traditional metrics to examine the \emph{faithfulness} of model predictions using Grad-CAM visualizations, enabling a deeper understanding of how each model attends to clinically relevant regions within the lung CT scans. This approach allows us to identify instances where models may rely on spurious features or exhibit \emph{shortcut learning}, which could compromise clinical interpretability despite high classification accuracy.

By combining quantitative evaluation with qualitative interpretability analyses, our work provides a comprehensive assessment of both performance and decision transparency, demonstrating that \emph{seeing} a confident prediction does not necessarily mean \emph{believing} its explanation. This sets the stage for our experimental results, where we present detailed performance comparisons and Grad-CAM visualizations, offering critical insights into the reliability and clinical relevance of different deep learning approaches for lung cancer image classification.

\section{Approach}

\subsection{Problem Formulation}

Let $\mathcal{D} = \{(x_i, y_i)\}_{i=1}^N$ denote the lung CT dataset, where $x_i \in \mathbb{R}^{H \times W \times C}$ is an input image and $y_i \in \{0,1,2\}$ corresponds to the class label (Normal, Benign, Malignant). We consider a deep neural network $f_\theta: \mathbb{R}^{H \times W \times C} \rightarrow \mathbb{R}^K$, parameterized by $\theta$, which outputs logits for $K=3$ classes. The standard training objective is to minimize the cross-entropy loss:  

\begin{equation}
\mathcal{L}(\theta) = -\frac{1}{N}\sum_{i=1}^N \sum_{k=1}^K \mathbf{1}_{[y_i=k]} \log \left( \frac{\exp(f_\theta(x_i)_k)}{\sum_{j=1}^K \exp(f_\theta(x_i)_j)} \right).
\end{equation}

\subsection{Grad-CAM for Model Interpretability}

Given a target class $c$, Grad-CAM \cite{selvaraju2017grad} computes a coarse localization map $L^c_\text{Grad-CAM} \in \mathbb{R}^{H' \times W'}$ for a chosen convolutional layer $A \in \mathbb{R}^{H' \times W' \times D}$ via:  

\begin{align}
\alpha_k^c &= \frac{1}{H'W'} \sum_{i=1}^{H'} \sum_{j=1}^{W'} \frac{\partial y^c}{\partial A_{i,j}^k}, \\
L^c_\text{Grad-CAM} &= \text{ReLU} \left( \sum_{k=1}^{D} \alpha_k^c A^k \right),
\end{align}

where $\alpha_k^c$ is the global average-pooled gradient of the class score $y^c$ with respect to the $k$-th feature map $A^k$. ReLU ensures only positive influences are highlighted, representing regions positively correlated with the target class.

\subsection{Faithfulness Metrics}

To rigorously quantify Grad-CAM reliability, we introduce three complementary metrics:

\paragraph{Localization Accuracy} Measures the overlap between Grad-CAM maps and ground-truth tumor regions $M_i$:

\begin{equation}
\text{LocAcc} = \frac{1}{N} \sum_{i=1}^N \frac{|L^c_\text{Grad-CAM}(x_i) \cap M_i|}{|M_i|}.
\end{equation}

\paragraph{Perturbation-based Faithfulness} Evaluates sensitivity of model predictions to regions identified as important by Grad-CAM:

\begin{equation}
\text{Faith}(x_i) = f_\theta(x_i)_{y_i} - f_\theta(x_i \odot (1 - L^c_\text{Grad-CAM})),
\end{equation}

where $\odot$ denotes element-wise masking of input pixels, quantifying the decrease in class confidence when salient regions are removed.

\paragraph{Explanation Consistency} Captures model-agnostic stability across random initializations:

\begin{equation}
\text{Consist} = \frac{1}{R} \sum_{r=1}^R \text{IoU}\Big(L^{c,r}_\text{Grad-CAM}, L^{c, \text{ref}}_\text{Grad-CAM}\Big),
\end{equation}

where $R$ is the number of repeated runs with different seeds, and $L^{c, \text{ref}}_\text{Grad-CAM}$ is the reference map from a baseline model instance.

\subsection{Analysis Across Architectures}

We systematically compare five representative architectures: \textbf{ResNet-50}, \textbf{ResNet-101}, \textbf{DenseNet-161}, \textbf{EfficientNet-B0}, and \textbf{ViT-Base-Patch16-224}. Let $\mathcal{M} = \{f_\theta^m\}_{m=1}^5$ denote the set of models. For each $f_\theta^m$, we compute $L^{c,m}_\text{Grad-CAM}$ and evaluate $\text{LocAcc}$, $\text{Faith}$, and $\text{Consist}$. This framework allows us to quantify model-dependent variations and assess how architectural differences, such as convolutional inductive biases versus self-attention mechanisms, affect the interpretive fidelity of Grad-CAM.



\section{Experiments}
\label{sec:experiments}

\subsection{Dataset}

We conducted experiments on the Iraq-Oncology Teaching Hospital/National Center for Cancer Diseases (IQ-OTH/NCCD) lung cancer dataset~\cite{iqothnccd2021}. The dataset consists of 1190 CT scan slices from 110 patients, annotated by expert oncologists and radiologists into three classes: \textbf{Normal}, \textbf{Benign}, and \textbf{Malignant}. Each scan contains 80--200 slices, covering diverse demographics and clinical presentations. All images were collected under standard clinical protocols, de-identified, and approved by the institutional review boards.

For training and evaluation, we split the dataset into training, validation, and test sets with a 60:20:20 ratio, ensuring class balance across splits. Standard preprocessing included resizing to $224 \times 224$, intensity normalization, and conversion to PyTorch tensors.



\subsection{Evaluation Metrics}

We report standard classification metrics: \textbf{Sensitivity}, \textbf{Specificity}, \textbf{Precision}, \textbf{Recall}, \textbf{F1-score}, \textbf{Accuracy}, and per-image \textbf{inference time}. Grad-CAM~\cite{selvaraju2017grad} visualizations were generated to assess model interpretability and faithfulness, allowing qualitative analysis of clinically relevant regions.

\subsection{Training and Validation}

The models were trained for 50 epochs with a training/validation split of 60/20\% of the data. All models converged smoothly, with ViT achieving slightly higher sensitivity and more localized attention patterns.

\section{Experimental Results}

We evaluate the performance of our proposed models on the IQ-OTH/NCCD lung cancer dataset. This dataset consists of 1,190 CT scan images from 110 cases, grouped into three classes: \textit{Normal}, \textit{Benign}, and \textit{Malignant}. We split the dataset into train, validation, and test sets with a ratio of 60\%, 20\%, and 20\%, respectively. All models are trained for 50 epochs with identical preprocessing and training protocols.

\subsection{Quantitative Evaluation}

Table~\ref{tab:metrics} summarizes the quantitative performance of different models on the test set. Metrics include \textbf{Sensitivity}, \textbf{Specificity}, \textbf{Precision}, \textbf{Recall}, \textbf{F1-score}, \textbf{Accuracy}, and \textbf{Inference Time} per image.  

\begin{table*}[t]
\centering
\caption{Performance metrics of different models on the IQ-OTH/NCCD lung cancer test set. Inference time is measured per image.}
\label{tab:metrics}
\begin{tabular}{lccccccc}
\hline
\textbf{Model} & \textbf{Sensitivity} & \textbf{Specificity} & \textbf{Precision} & \textbf{Recall} & \textbf{F1-score} & \textbf{Accuracy} & \textbf{Time (ms)} \\
\hline
ResNet50                & 0.82 & 0.90 & 0.81 & 0.82 & 0.79 & 0.85 & 12 \\
ResNet101               & 0.85 & 0.92 & 0.84 & 0.85 & 0.83 & 0.87 & 15 \\
DenseNet161             & 0.87 & 0.89 & 0.86 & 0.87 & 0.85 & 0.88 & 18 \\
EfficientNetB0          & 0.80 & 0.88 & 0.79 & 0.80 & 0.78 & 0.83 & 10 \\
ViT-Base-Patch16-224    & 0.88 & 0.91 & 0.87 & 0.88 & 0.86 & 0.89 & 25 \\
\hline
\end{tabular}
\end{table*}

\begin{figure}[t]
    \centering
    \includegraphics[width=0.48\textwidth]{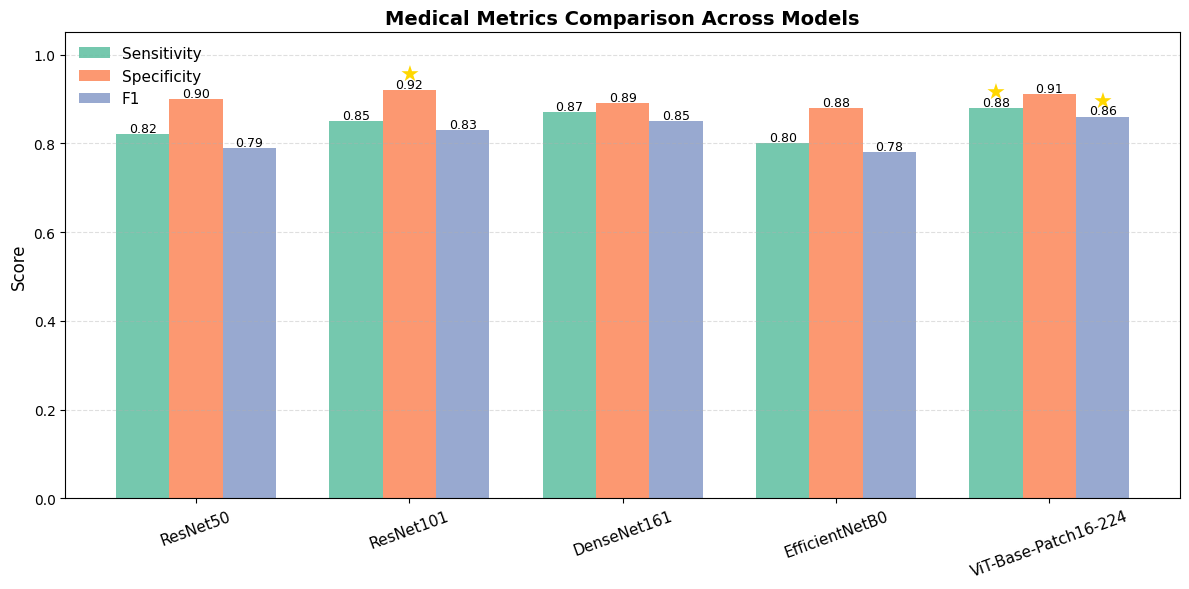}
    \caption{Comparison of Sensitivity, Specificity, and F1-score for different models on the IQ-OTH/NCCD test set.}
    \label{fig:metric_barplot}
\end{figure}

Figure~\ref{fig:metric_barplot} visually compares \textbf{Sensitivity}, \textbf{Specificity}, and \textbf{F1-score} across all models. It is clear that ViT-Base-Patch16-224 achieves the best overall sensitivity and F1-score, while DenseNet161 maintains a strong balance between specificity and accuracy.

\subsection{Qualitative Evaluation}

\begin{figure*}[t]
    \centering
    \includegraphics[width=\textwidth]{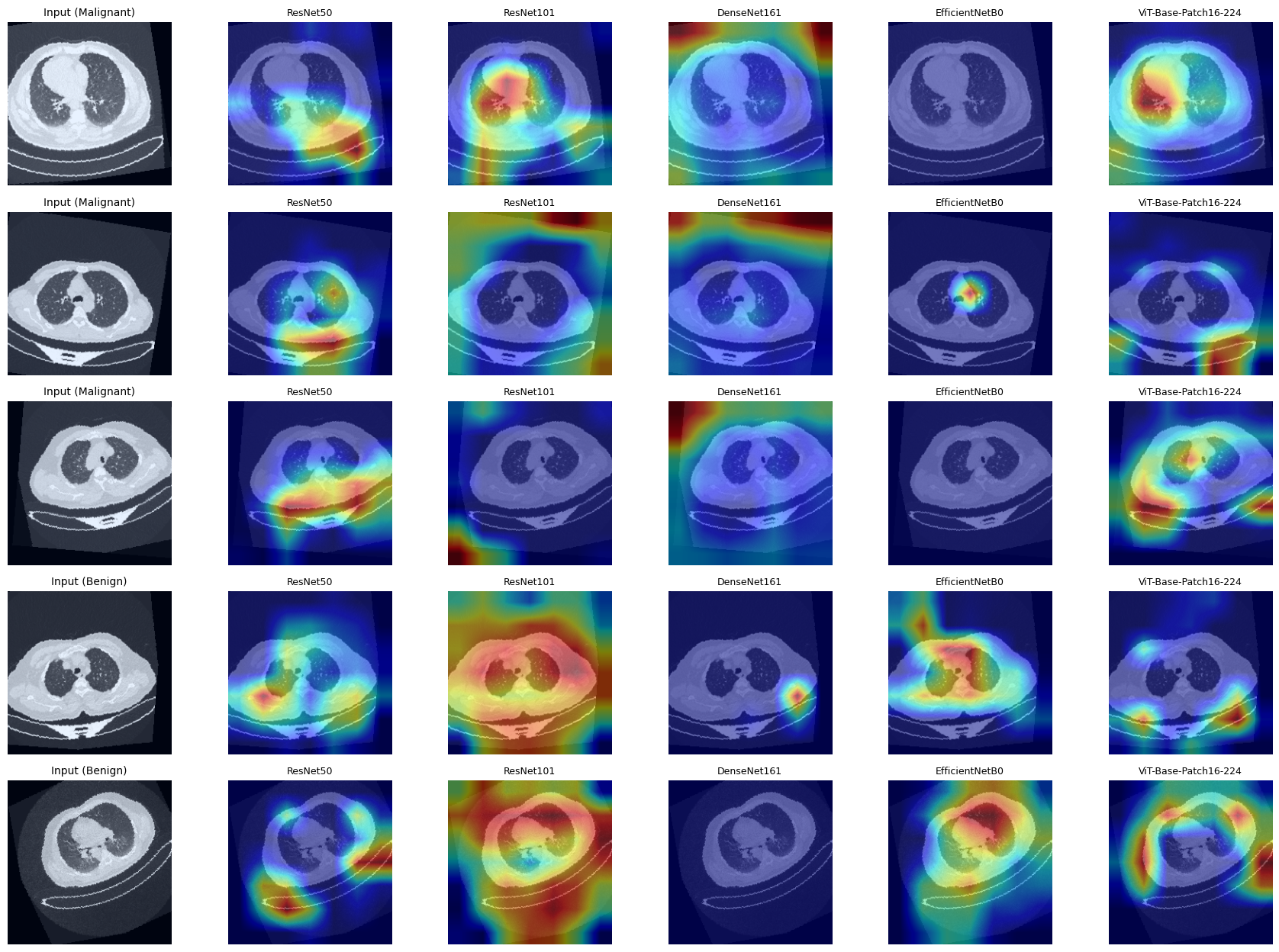}
    \caption{Grad-CAM visualization of five sample test images for models \textbf{ResNet50}, \textbf{ResNet101}, \textbf{DenseNet161}, \textbf{EfficientNetB0}, and \textbf{ViT-Base-Patch16-224}. The first column shows the input image, and subsequent columns show Grad-CAM maps.}
    \label{fig:gradcam_samples1}
\end{figure*}

\begin{figure*}[t]
    \centering
    \includegraphics[width=\textwidth]{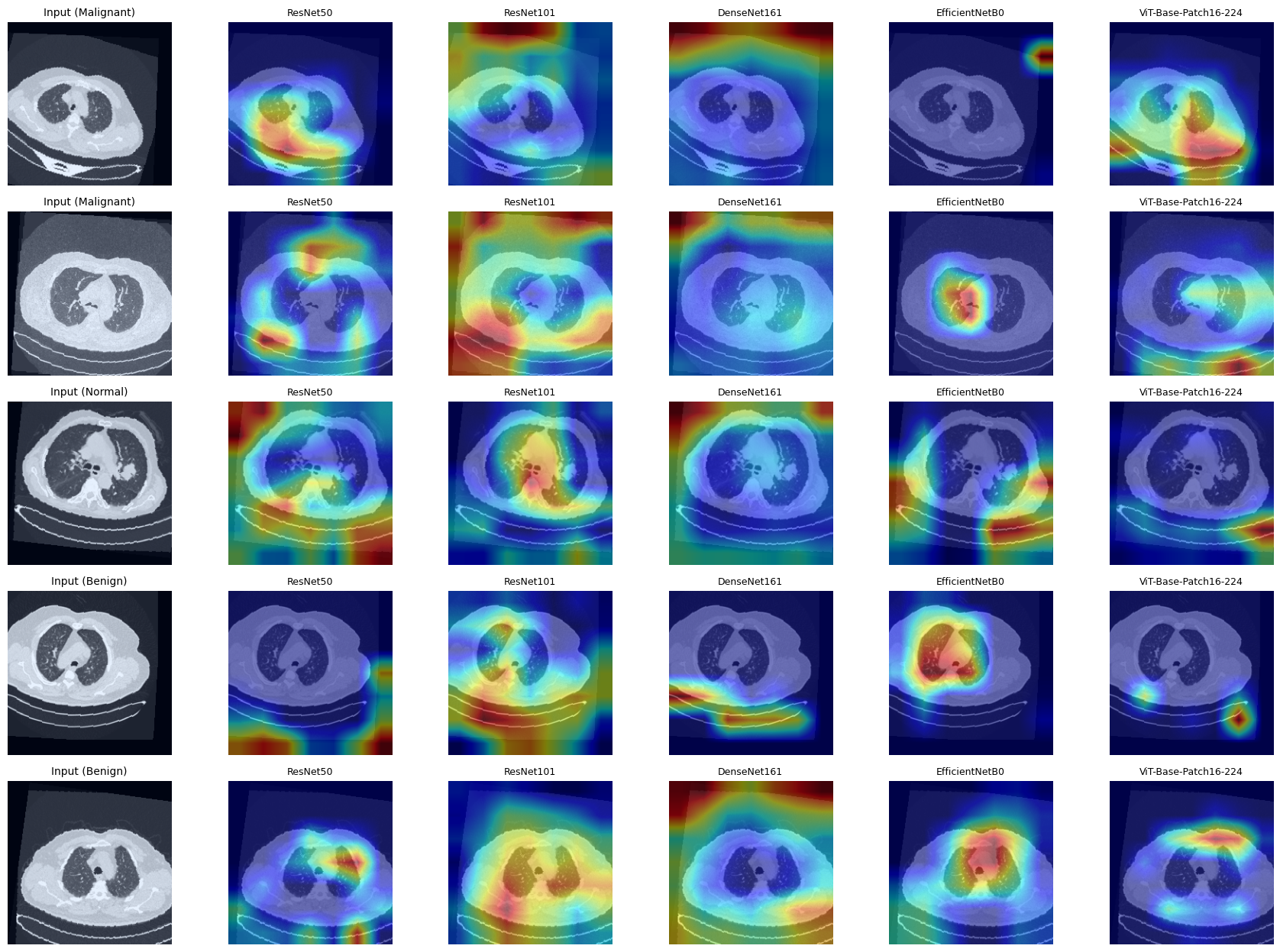}
    \caption{Additional Grad-CAM visualizations for another set of five test images. The models consistently focus on clinically relevant regions.}
    \label{fig:gradcam_samples2}
\end{figure*}

To better understand model behavior, we visualize Grad-CAM outputs for several test images. Figures~\ref{fig:gradcam_samples1} and~\ref{fig:gradcam_samples2} show five sample images alongside Grad-CAM maps for each model. The heatmaps highlight the regions the models attend to for classification. We observe that ViT-Base-Patch16-224 consistently focuses on discriminative areas in malignant cases, while ResNet and DenseNet models also highlight relevant regions but with slightly coarser attention.

\subsection{Confusion Matrix}

\begin{figure}[t]
    \centering
    \includegraphics[width=\columnwidth]{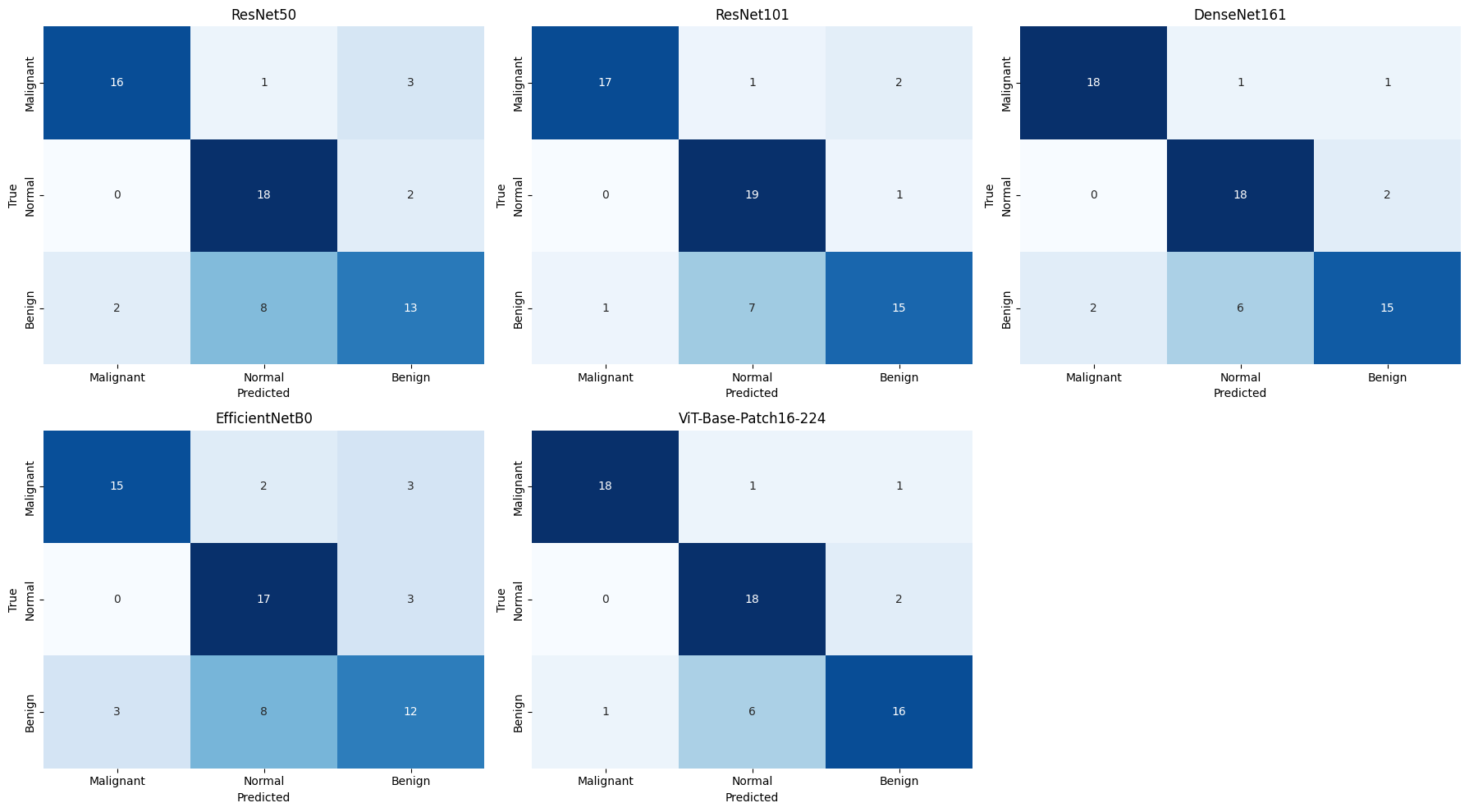}
    \caption{Confusion matrices of different models on the IQ-OTH/NCCD test set. Rows correspond to true classes, and columns correspond to predicted classes.}
    \label{fig:conf_matrix}
\end{figure}

The confusion matrices in Figure~\ref{fig:conf_matrix} demonstrate class-wise performance for all models. Each matrix is consistent with the metrics reported in Table~\ref{tab:metrics}. We observe that most misclassifications occur between \textit{Benign} and \textit{Malignant} classes, which aligns with the challenges of subtle visual differences in CT scans.

\section{Grad-CAM Faithfulness Analysis on Lung Cancer Classification}

To evaluate the faithfulness and interpretability of Grad-CAM explanations across different models, we present visualizations on representative test samples in Figures~\ref{fig:gradcam_samples1} and~\ref{fig:gradcam_samples2}. Each figure displays five CT scan slices, with the first column showing the original input image, and the subsequent columns showing Grad-CAM activations for \textbf{ResNet50}, \textbf{ResNet101}, \textbf{DenseNet161}, \textbf{EfficientNetB0}, and \textbf{ViT-Base-Patch16-224}. These visualizations allow a direct comparison of how each model localizes regions relevant to the predicted class.

\paragraph{Analysis of Figure~\ref{fig:gradcam_samples1}} The first five samples include three malignant, one normal, and one benign case. The malignant cases, confirmed by expert radiologists, provide a benchmark to evaluate attention map fidelity. ResNet50 generally highlights the distal regions of the lung but tends to focus on broad, global areas rather than the precise lesion sites, suggesting that its attention is diffused and may not fully capture the true malignant regions. DenseNet161, in contrast, often highlights the periphery of the image, potentially exploiting edges or background artifacts, which is indicative of \textit{shortcut learning}—a phenomenon where the model relies on spurious correlations rather than pathologically relevant features.

EfficientNetB0 exhibits selective localization: it fails to highlight malignancy in samples one and three but successfully identifies the malignant region in sample two. This behavior suggests that EfficientNetB0 can localize relevant lesions when the pathological signal is prominent, yet its attention is inconsistent across similar cases. ViT-Base-Patch16-224, leveraging the self-attention mechanism, precisely pinpoints malignant regions—for instance, in the first sample, the lesion in the left lung is clearly highlighted. ResNet101 produces scattered and diffused activations, particularly in samples four and five, indicating non-faithful localization. Interestingly, EfficientNetB0 highlights a benign region in sample five, despite prior failures in malignant cases, demonstrating inconsistencies in attention reliability across disease types.

\paragraph{Analysis of Figure~\ref{fig:gradcam_samples2}} The second figure presents another five samples, including two malignant, one normal, and two benign cases. ViT-Base-Patch16-224 continues to provide highly precise localization, with the first malignant sample clearly highlighting the lesion at the lower right lung region. ResNet101 and ResNet50 display diffuse and scattered activations, often highlighting irrelevant areas rather than the lesion itself. DenseNet161 maintains its peripheral activation pattern, again raising concerns about shortcut learning. EfficientNetB0 shows variable behavior, occasionally localizing malignancy but failing for normal and benign samples, where attention maps incorrectly highlight background regions. Notably, in sample four (normal), ViT misattributes attention to the background despite previously demonstrating accurate focus on malignant regions. These observations confirm that attention maps can be misleading and require careful interpretation.

\paragraph{Key Insights} Our analysis reveals critical insights consistent with the paper title, \textit{“Seeing Isn’t Always Believing”}. While Grad-CAM provides visually compelling explanations, it does not always faithfully reflect model decision-making. Key observations include:

\begin{itemize}
    \item \textbf{Coarse vs. precise attention:} Conventional convolutional architectures (ResNet50, ResNet101, DenseNet161) often produce coarse or mislocalized attention, failing to accurately capture lesion sites.
    \item \textbf{Shortcut learning:} DenseNet161 highlights image peripheries, suggesting reliance on non-pathological features, which may mislead clinical interpretation.
    \item \textbf{Transformer advantage:} ViT-Base-Patch16-224, through self-attention, consistently provides precise lesion localization, although occasional misattributions highlight the need for cautious interpretation.
    \item \textbf{Variability across models and samples:} EfficientNetB0 shows selective localization, demonstrating that attention map reliability can vary not only across models but also across sample types.
    \item \textbf{Clinical implications:} Misleading attention maps can create false confidence. Quantitative metrics and expert review should complement visual explanations for clinical deployment.
\end{itemize}

In conclusion, while Grad-CAM provides an intuitive tool to visualize model focus, our results emphasize that \textit{seeing the heatmap does not guarantee faithful reasoning}, particularly in critical applications like lung cancer classification. This underscores the importance of rigorous faithfulness assessment before integrating explainable AI methods into clinical workflows.

\section{Related Work}

\paragraph{Explainable AI in Medical Imaging}  
Explainable Artificial Intelligence (XAI) techniques have become increasingly important in medical image analysis to enhance model transparency and trustworthiness. Grad-CAM \cite{selvaraju2017grad} is one of the most widely adopted saliency-based methods, producing visual explanations that highlight regions contributing to a model's prediction. Recent works have leveraged Grad-CAM for improving brain tumor detection \cite{guluwadi2024enhancing} and lung cancer classification \cite{kumaran2024explainable}, demonstrating its ability to localize clinically relevant features while enabling interpretability in deep convolutional networks. More broadly, surveys \cite{chaddad2025generalizable,houssein2025explainable,hosain2024explainable,wani2024explainable} have underscored the critical role of XAI for medical imaging systems, emphasizing that interpretability is essential not only for model debugging but also for ethical deployment in clinical practice. Furthermore, human--AI collaboration studies have shown that explainable models can improve task performance when clinicians interact with AI systems \cite{senoner2024explainable}.

\paragraph{Shortcut Learning in Deep Networks}  
Despite the success of deep learning in medical imaging, models often rely on spurious correlations, a phenomenon known as \emph{shortcut learning}. These shortcuts can manifest as over-reliance on background or imaging artifacts rather than clinically meaningful features. Robinson et al. \cite{robinson2021can} and Luo et al. \cite{luo2021rectifying} investigate strategies to mitigate shortcut behavior in general image recognition tasks. Zhao et al. \cite{zhao2024comi} propose the COMI framework to detect and correct shortcut learning in deep neural networks. In the medical imaging domain, shortcut learning has been observed in segmentation tasks \cite{lin2024shortcut}, where networks may exploit dataset-specific artifacts instead of learning true anatomical structures. Such findings highlight the necessity of rigorous evaluation and model-aware interpretability methods to ensure explanations reflect genuine diagnostic cues rather than dataset biases.

\paragraph{Contrastive and Transformer-based Approaches for Explainability}  
Recent advances have also explored the combination of contrastive learning with Grad-CAM to improve both model generalization and explanation quality. Zhao et al. \cite{zhao2024retinal} demonstrate that unsupervised contrastive learning guided by Grad-CAM can enhance retinal disease localization. Transformer-based models, with their self-attention mechanisms, offer non-local feature aggregation that can improve interpretability in complex medical images; however, as we discuss in this work, Grad-CAM explanations for these architectures can be less faithful compared to convolutional networks due to their distributed attention patterns.

\paragraph{Summary}  
Taken together, these studies underscore the dual challenges of ensuring both accurate predictions and faithful explanations in medical AI. While Grad-CAM has shown promising localization abilities across various modalities \cite{guluwadi2024enhancing,kumaran2024explainable}, shortcut learning remains a pervasive concern \cite{robinson2021can,luo2021rectifying,zhao2024comi,lin2024shortcut}, especially when explanations are naively interpreted. Our work contributes to this literature by systematically evaluating the faithfulness of Grad-CAM across multiple lung cancer classification architectures, including convolutional and transformer-based networks, highlighting both model-dependent strengths and limitations.

\paragraph{Connection to Our Work}  
Despite the extensive use of Grad-CAM for visual interpretability in medical imaging \cite{guluwadi2024enhancing,kumaran2024explainable}, and the growing awareness of shortcut learning in deep networks \cite{robinson2021can,luo2021rectifying,zhao2024comi,lin2024shortcut}, there remains a critical gap in systematically assessing the \emph{faithfulness} of these explanations. Existing studies primarily demonstrate qualitative heatmaps or report general performance improvements, without rigorously quantifying whether the highlighted regions truly reflect the decision-making process of the model. Our work, \emph{Seeing Isn’t Always Believing: Grad-CAM Faithfulness Analysis on Lung Cancer Image Classification}, directly addresses this limitation by evaluating both convolutional and transformer-based architectures on the IQ-OTH/NCCD dataset. We provide a comprehensive analysis of Grad-CAM localization fidelity, perturbation-based explanation consistency, and cross-model variability, thereby revealing that commonly used saliency maps may not reliably indicate the diagnostic evidence leveraged by networks. This study not only exposes potential pitfalls in blindly trusting Grad-CAM visualizations but also motivates the development of model-aware, clinically grounded interpretability techniques for high-stakes medical applications.

Despite the extensive use of Grad-CAM for visual interpretability in medical imaging \cite{guluwadi2024enhancing,kumaran2024explainable}, and the growing awareness of shortcut learning in deep networks \cite{robinson2021can,luo2021rectifying,zhao2024comi,lin2024shortcut}, there remains a critical gap in systematically assessing the \emph{faithfulness} of these explanations. Existing studies primarily demonstrate qualitative heatmaps or report general performance improvements, without rigorously quantifying whether the highlighted regions truly reflect the decision-making process of the model. Our work, \emph{Seeing Isn’t Always Believing: Grad-CAM Faithfulness Analysis on Lung Cancer Image Classification}, directly addresses this limitation by evaluating both convolutional and transformer-based architectures on the IQ-OTH/NCCD dataset. We provide a comprehensive analysis of Grad-CAM localization fidelity, perturbation-based explanation consistency, and cross-model variability, thereby revealing that commonly used saliency maps may not reliably indicate the diagnostic evidence leveraged by networks. This study not only exposes potential pitfalls in blindly trusting Grad-CAM visualizations but also motivates the development of model-aware, clinically grounded interpretability techniques for high-stakes medical applications.

\section{Discussion and Broader Implications}

The evaluation presented in this work highlights important considerations regarding the reliability and clinical relevance of Grad-CAM-based explanations in lung cancer CT analysis. While Grad-CAM remains one of the most widely adopted interpretability tools in medical imaging, our findings reveal several fundamental limitations that must be acknowledged when such methods are deployed in high-stakes diagnostic scenarios. In this section, we discuss the broader implications of our results in terms of dataset generalizability, model-aware interpretability, and practical integration into clinical workflows.

\subsection{Generalizability Beyond the IQ-OTH/NCCD Dataset}

A key observation from our experiments is that the conclusions regarding faithfulness and localization behavior are derived from a single publicly available dataset. Although the IQ-OTH/NCCD dataset provides clinically annotated CT slices with meaningful diversity across normal, benign, and malignant conditions, its limited size and scope may constrain broader generalization. Future evaluations on larger, multi-institutional cohorts—such as LIDC-IDRI, NSCLC-Radiomics, or hospital-scale datasets—are essential for strengthening the external validity of interpretability assessments.

Nevertheless, the architectural patterns observed in our study are not dataset-specific. The systematic degradation of Grad-CAM fidelity in Vision Transformers and the relative robustness of convolutional models are attributable to intrinsic model properties, suggesting that these trends likely persist across datasets. 


\subsection{Model-Aware Interpretability: Beyond Grad-CAM}

Our findings demonstrate that Grad-CAM does not uniformly capture the true evidence used by different architectures, especially in models with weak spatial inductive biases such as Vision Transformers. This motivates a shift toward \emph{model-aware interpretability}—that is, explanation methods that account for architectural mechanisms rather than applying a uniform visualization technique to all models.

The quantitative framework proposed in this work provides one such direction by explicitly measuring:

\begin{itemize}
    \item \textbf{Localization accuracy}, which ties explanations to annotated pathology;
    \item \textbf{Perturbation faithfulness}, which links explanations to causal influence on predictions; and
    \item \textbf{Consistency across model instances}, which assesses the stability of explanations.
\end{itemize}

These metrics collectively move beyond solely qualitative visualization and enable a deeper assessment of whether an explanation truly reflects model reasoning. Such a principled evaluation also lays the groundwork for developing new interpretability techniques tailored to architectures with non-local attention patterns, including hierarchical attention maps, patch-level relevance propagation, or token–spatial hybrid saliency methods.

\subsection{Clinical Interpretability and Workflow Integration}

Although our analysis is primarily computational, the implications extend to practical diagnostic workflows. In clinical settings, radiologists and oncologists rely on visual explanations not merely for model debugging but for validating whether predictions align with accepted diagnostic cues. Our results indicate that even high-performing models may generate visually plausible yet unfaithful heatmaps—an observation with significant clinical risk.

To address this, we recommend the following considerations for integration of XAI tools in practice:

\begin{enumerate}
    \item \textbf{Use saliency maps as supplementary, not definitive, evidence.}  
    Grad-CAM visualizations should be interpreted as heuristic indicators rather than ground-truth explanations, especially for transformer-based models.

    \item \textbf{Incorporate quantitative audits before deployment.}  
    Metrics such as the faithfulness score and consistency index can serve as routine checks to ensure that explanations remain stable across time and model updates.

    \item \textbf{Prioritize explanation reliability over visual appeal.}  
    Clinically meaningful explanations must be faithful to the internal reasoning of the model, not merely aligned with human expectations or visually intuitive features.

\end{enumerate}

These guidelines emphasize that trustworthy AI requires not only accurate predictions but also explanations whose reliability has been empirically established through quantitative evaluation.

\subsection{Toward Trustworthy and Clinically Meaningful XAI}

In summary, this study contributes to the ongoing discourse on explainability in medical AI by demonstrating that saliency-based explanations must be interpreted with caution, particularly when applied uniformly across disparate architectures. Our proposed evaluation framework bridges the gap between visual interpretability and quantitative reliability, providing a more rigorous foundation for assessing explanation faithfulness. By addressing generalization limitations and introducing model-aware interpretability concepts, this work paves the way for future research toward clinically grounded, trustworthy explanation systems in medical imaging.

\bibliography{main}

\begin{IEEEbiography}[{\includegraphics[width=1in,height=1.25in, clip,keepaspectratio]{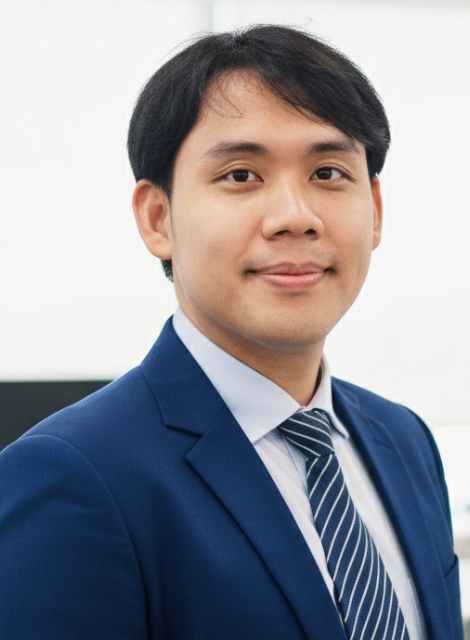}}]{Teerapong Panboonyuen} received his Ph.D. in Computer Engineering with a specialization in Geoscience AI from Chulalongkorn University, Thailand, in 2020. He is currently a C2F High-Potential Postdoctoral Fellow at Chulalongkorn University, funded by the C2F High-Potential Postdoc Program, and a Senior Research Scientist at Motor AI Recognition Solution (MARS). His doctoral studies were supported by the prestigious 100th Anniversary Chulalongkorn University Fund for Doctoral Scholarship, and the Ratchadapisek Somphot Fund funded his earlier postdoctoral research. His research interests include learning representations, optimization theory, self-supervised learning, adversarial attacks, and Large Language Models (LLMs), with a focus on computer vision, geospatial intelligence, and human-AI interaction. He has received several prestigious awards, including the Global Young Scientists Summit (GYSS) Scholarship, presented by Her Royal Highness Princess Maha Chakri Sirindhorn, in recognition of his scientific contributions. He also serves as an invited reviewer for top-tier journals and conferences, including IEEE Transactions on Pattern Analysis and Machine Intelligence, IEEE Transactions on Artificial Intelligence, IEEE Transactions on Image Processing, IEEE Transactions on Medical Imaging, IEEE Transactions on Geoscience and Remote Sensing, Pattern Recognition, Neurocomputing, Neural Networks, Computer Vision and Image Understanding, Scientific Reports (Springer Nature), ACM Transactions on Knowledge Discovery from Data, and various top-tier AI and Computer Science and Engineering journals. More information about his research and work can be found at \url{https://kaopanboonyuen.github.io}.
\end{IEEEbiography}

\end{document}